\def\year{2020}\relax
\documentclass[letterpaper]{article} % DO NOT CHANGE THIS
\usepackage{aaai20}  % DO NOT CHANGE THIS
\usepackage{times}  % DO NOT CHANGE THIS
\usepackage{helvet} % DO NOT CHANGE THIS
\usepackage{courier}  % DO NOT CHANGE THIS
\usepackage[hyphens]{url}  % DO NOT CHANGE THIS
\usepackage{graphicx} % DO NOT CHANGE THIS
\usepackage{graphicx}  % DO NOT CHANGE THIS
\newcommand{\citet}[1]{\citeauthor{#1} \shortcite{#1}}
\newcommand{\citep}{\cite}

\usepackage{soul}
\usepackage{subfig}
\usepackage{booktabs}
\usepackage{bm}
\usepackage{amssymb}
\usepackage{amsmath}
\DeclareMathOperator*{\argmax}{arg\,max}
\DeclareMathOperator*{\softmax}{softmax}
\DeclareMathOperator*{\abs}{abs}

\DeclareMathOperator*{\transfer}{Transfer}
\DeclareMathOperator*{\fsmt}{FSMT}
\DeclareMathOperator*{\ced}{CED}
\newcommand{\vb}{\bm{b}}
\newcommand{\vh}{\bm{h}}
\newcommand{\vW}{\bm{W}}
\newcommand{\vX}{\bm{X}}
\newcommand{\vY}{\bm{Y}}
\newcommand{\vtheta}{\bm{\theta}}
\newcommand{\logll}{\mathcal{L}}
\newcommand{\bftab}{\fontseries{b}\selectfont}

\usepackage{xspace}
\newcommand{\methodY}{Online Target Inference\xspace}
\newcommand{\methodEll}{Online Style Inference\xspace}
\newcommand{\baseDS}{NMT DS-Tag\xspace}
\newcommand{\baseMTL}{Multi-Task\xspace}
\newcommand{\baseMTLDS}{Multi-Task DS-Tag\xspace}

\title{Controlling Neural Machine Translation Formality with Synthetic Supervision}
\author{
	Xing Niu,\textsuperscript{\rm 1}
	Marine Carpuat\textsuperscript{\rm 2}\\
	\textsuperscript{\rm 1}Amazon AWS AI,
	\textsuperscript{\rm 2}University of Maryland\\
	xingniu@amazon.com, marine@cs.umd.edu
}

\begin{document}

\maketitle

\begin{abstract}
This work aims to produce translations that convey source language content at a formality level that is appropriate for a particular audience. Framing this problem as a neural sequence-to-sequence task ideally requires training triplets consisting of a bilingual sentence pair labeled with target language formality. However, in practice, available training examples are limited to English sentence pairs of different styles, and bilingual parallel sentences of unknown formality. We introduce a novel training scheme for multi-task models that automatically generates synthetic training triplets by inferring the missing element on the fly, thus enabling end-to-end training. Comprehensive automatic and human assessments show that our best model outperforms existing models by producing translations that better match desired formality levels while preserving the source meaning.\footnote{This work was done when the first author was at the University of Maryland.}
\end{abstract}

\section{Introduction}

Producing language in the appropriate style is a requirement for natural language generation, as the style of a text conveys information beyond its literal meaning \citep{Hovy87}. This also applies to translation: professional translators adapt translations to their audience \citep{NidaT03}, yet the output style has been overlooked in machine translation. For example, the French sentence ``Bonne id\'ee, mais elle ne convient pas ici." could be translated to ``Good idea, but it doesn't fit here.", which is informal because it elides the subject, uses contractions and chained clauses. It could also be translated more formally to ``This is a helpful idea. However, it is not suitable for this purpose.", which is grammatically complete and uses more formal and precise terms.

We recently addressed this gap by introducing the task of Formality-Sensitive Machine Translation (FSMT), where given a French sentence and a desired formality level, systems are asked to produce an English translation at the specified formality level \citep{NiuMC17}. Building FSMT systems is challenging because of the lack of appropriate training data: bilingual parallel corpora do not come with formality annotations, and parallel corpora of a given provenance do not have a uniform style. Previously, we took a multi-task approach based on sequence-to-sequence models that were trained to perform French-English translation and English formality transfer \citep{RaoT18} jointly \citep{NiuRC18}. The resulting multi-task model performs zero-shot FSMT as it has never been exposed to training samples annotated with both reference translation and formality labels.

In this work, we hypothesize that exposing multi-task models to training samples that directly match the FSMT task can help generate formal and informal outputs that differ from each other, and where formality rewrites do not introduce translation errors. We introduce \methodEll, an approach to simulate direct supervision by predicting the target formality of parallel sentence pairs on the fly at training time, thus enabling end-to-end training. We also present a variant of side constraints \cite{SennrichHB16} that improves formality control given inputs of arbitrary formality level.\footnote{Source code: https://github.com/xingniu/multitask-ft-fsmt.}

We conduct a comprehensive automatic and human evaluation of the resulting FSMT systems. First, we show that \methodEll introduces more differences between formal and informal translations of the same input, using automatic metrics to quantify lexical and positional differences. Second, we conduct a human evaluation which shows that \methodEll preserves the meaning of the input and introduces stronger formality differences compared to a strong baseline. Finally, we analyze the diversity of transformations between formal and informal outputs produced by our approach.

\section{A Neural FSMT Model}
\label{sec:fsmt}

Neural Machine Translation (NMT) models compute the conditional probability $P(\vY|\vX)$ of translating a source sentence, $\vX=(x_1,\dots,x_n)$, to a target sentence, $\vY=(y_1,\dots,y_m)$. By contrast, FSMT requires producing the most likely translation at the given formality level $\ell$:
\begin{equation}\label{eq:fsmt}
\hat{\vY} = \argmax_{\vY_\ell} P(\vY_\ell\,|\,\vX,\ell;\vtheta).
\end{equation}
Ideally, the FSMT model should be trained on triplets $(\vX,\ell,\vY_\ell)_{1\dots N}$, but in practice, such training data is not easy to acquire. We tackle this problem by training a cross-lingual machine translation model (French$\rightarrow$English) and a monolingual bidirectional formality transfer model (Formal-English$\leftrightarrow$Informal-English) jointly \citep{NiuRC18}. Specifically, the model is trained on the combination of $(\vX,\vY)_{1\dots N_1}$ and $(\vY_{\bar{\ell}},\ell,\vY_\ell)_{1\dots N_2}$, where $\vY_{\bar{\ell}}$ and $\vY_\ell$ have opposite formality levels. The joint model is able to perform zero-shot FSMT by optimizing $\logll_{MT}+\logll_{FT}$, where
\begin{align}\label{eq:fsmt_obj}
\logll_{MT} &= \sum_{(\vX,\vY)} \log P(\vY|\vX;\vtheta), \\
\logll_{FT} &= \sum_{(\vY_{\bar{\ell}},\ell,\vY_\ell)} \log P(\vY_\ell\,|\,\vY_{\bar{\ell}},\ell;\vtheta).
\end{align}

\subsection{Controlling Output Language Formality}

FSMT shares the goal of producing output sentences of a given formality with monolingual formality style transfer tasks. In both cases, the source sentence usually carries its own style and the model should be able to override it with the independent style $\ell$. Previously, we achieved this using an attentional sequence-to-sequence model with side constraints \citep{SennrichHB16}, i.e., attaching a style tag (e.g., \texttt{<2Formal>}) to the beginning of each source example \citep{NiuRC18}. In this work, similar to \citet{WangZZXZ18} and \citet{LampleSSDRB19}, we attach style tags to both source and target sequences to better control output formality given inputs of arbitrary style.

\citet{SennrichHB16} hypothesize that source-side tags control the target style because the model ``learns to pay attention to the side constraints", but it has not been verified empirically. We hypothesize that the source style tag also influences the encoder hidden states, and providing a target-side tag lets the decoder benefit from encoding style more directly. This approach yields a single model which is able to both transfer formality (e.g., from formal to informal, or vice versa) and preserve formality (e.g., producing an informal output given an informal input).

\subsection{Synthetic Supervision --- \methodEll}
\label{sec:synthetic}

Prior work on multilingual NMT shows that the translation quality on zero-shot tasks often significantly lags behind when supervision is provided \cite{JohnsonSLKWCTVW17}. We address this problem by simulating the supervision, i.e., generating synthetic training triplets $(\vX,\ell,\vY)$ by using the FSMT model itself to predict the missing element of the triplet from parallel sentence pairs $(\vX,\vY)$.

We introduce \textbf{\methodEll (OSI)} to generate synthetic triplets. Given a translation example $(\vX,\vY)$, we view predicting the formality of $\vY$, i.e., $\ell_{\vY}$, as unsupervised classification using only the pre-trained FSMT model. 

\begin{figure}[t]
	\centering
	\includegraphics[width=\columnwidth]{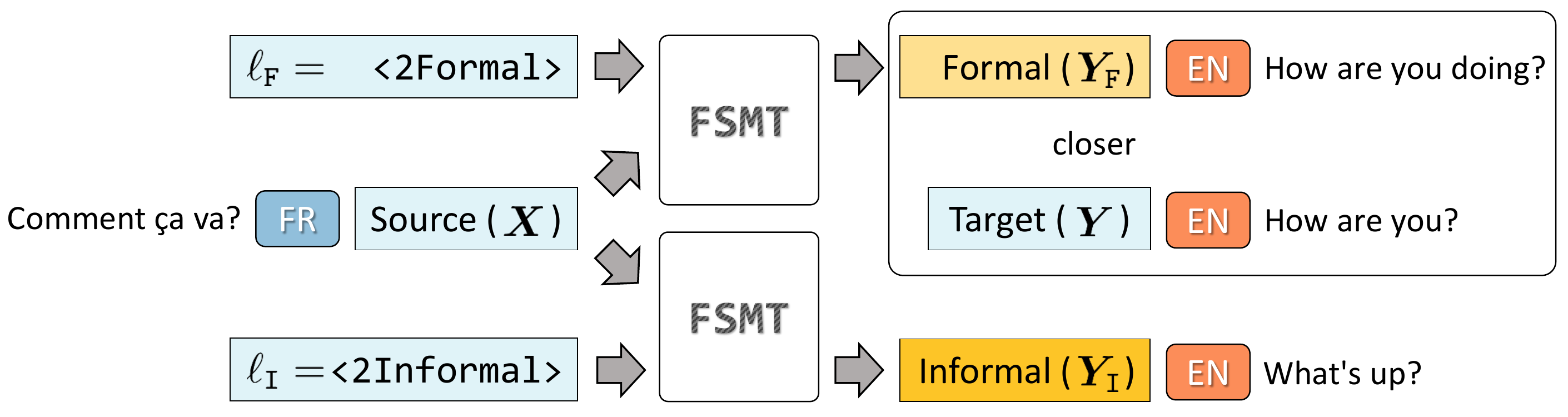}
	\caption{\methodEll. Given a translation example $(\vX,\vY)$, FSMT produces both informal and formal translations of $\vX$, i.e., $\vY_\texttt{I}=\fsmt(\vX,\ell_\texttt{I})$ and  $\vY_\texttt{F}=\fsmt(\vX,\ell_\texttt{F})$. $\vY$ is labeled as formal since it is closer to $\vY_\texttt{F}$ than $\vY_\texttt{I}$.}
	\label{fig:osi}
\end{figure}

As illustrated in Figure~\ref{fig:osi}, we use FSMT to produce both informal and formal translations of the same input, $\vY_\texttt{I}=\fsmt(\vX,\ell_\texttt{I})$ and  $\vY_\texttt{F}=\fsmt(\vX,\ell_\texttt{F})$ respectively.\footnote{$\vY_\texttt{I}$ and $\vY_\texttt{F}$ are generated with the \textit{teacher forcing} strategy \citep{WilliamsZ89} given the ground-truth $\vY$.} We hypothesize that the style of the reference translation $\vY$ can be predicted based on its distance from these two translations. For example, if $\vY$ is formal, it should be closer to $\vY_\texttt{F}$ than $\vY_\texttt{I}$. We measure the closeness by cross-entropy difference \citep[CED]{MooreL10}, i.e., we calculate the difference of their per-token cross-entropy scores, $\ced(\vY_\texttt{I},\vY_\texttt{F})=H_{\vY}(\vY_\texttt{I})-H_{\vY}(\vY_\texttt{F})$. The larger it is, the closer $\vY$ is to $\vY_\texttt{F}$.

Given a positive threshold $\tau$, we label $\ell_{\vY}=\texttt{<2Informal>}$ if $\ced(\vY_\texttt{I},\vY_\texttt{F})<-\tau$, label $\ell_{\vY}=\texttt{<2Formal>}$ if $\ced(\vY_\texttt{I},\vY_\texttt{F})>\tau$, and label $\ell_{\vY}=\texttt{<2Unknown>}$ otherwise. The threshold $\tau$ is chosen dynamically for each mini-batch, and it is equal to the mean of absolute token-level CED of all tokens within a mini-batch. Finally, we are able to generate a synthetic training sample, $(\vX,\ell_{\vY},\vY)$, on the fly and optimize $\logll_{FT}+\logll_{OSI}$, where
\begin{equation}\label{eq:osi_obj}
\logll_{OSI} = \sum_{(\vX,\ell_{\vY},\vY)} \log P(\vY\,|\,\vX,\ell_{\vY};\vtheta).
\end{equation}

Instead of inferring the style label $\ell$, we could obtain synthetic training triplets by generating target sequences for a desired $\ell$ and a given input $\vX$. We experiment with one such approach, which we call \textbf{\methodY (OTI)}, and will compare it with OSI empirically for completeness. However, OTI is expected to be less effective as generates complete output sequences and is therefore more likely to introduce noise in synthetic examples. Given the bilingual parallel sentence pair ($\vX$, $\vY$) and a randomly selected target formality $\ell$ from $\{\texttt{<2Informal>},\texttt{<2Formal>}\}$, we could use the FSMT model to produce a formality-constrained translation $\vY_\ell^1=\fsmt(\vX,\ell)$. We estimate the quality of $\vY_\ell^1$ indirectly using the multi-task nature of the FSMT models. The FSMT model can also manipulate the formality level of the target side $\vY$ via monolingual formality transfer to produce $\vY_\ell^2=\transfer(\vY,\ell)$. We hypothesize that the predictions made by these two different paths should be consistent.

The quality of $\vY_\ell^2$ is presumably more reliable than $\vY_\ell^1$, because the embedded transfer model is trained with direct supervision. We empirically get $\vY_\ell^2$ via greedy search on the fly during the training and use it as the label. Finally, we optimize $\logll_{MT}+\logll_{FT}+\alpha\logll_{OTI}$, where
\begin{equation}\label{eq:oti_obj}
\logll_{OTI} = \sum_{(\vX,\ell,\vY_\ell^2)} \log P(\vY_\ell^2\,|\,\vX,\ell;\vtheta).
\end{equation}

\section{Experimental Set-Up}

We design experiments to evaluate the impact of our approaches to (1) formality control, and (2) synthetic supervision. We first evaluate formality control on an English style transfer task which provides multiple reference translations to reliably evaluate formality transfer with automatic metrics. We then quantify the differences between formal and informal FSMT translation when using synthetic supervision. Finally, we design a manual evaluation to assess whether synthetic supervision improves over multi-task FSMT. All these experiments share the following set-up.

\begin{table}[t]
	\centering
	\begin{tabular}{l|r|r}
		Corpus & \# sentences & \# EN tokens \\ % # tokens (FR)
		\hline
		Europarl.v7 & 1,670,324 & 39,789,959 \\ % 43,614,024
		News-Commentary.v14 & 276,358 & 6,386,435 \\ % 7,549,393
		OpenSubtitles2016 & 16,000,000 & 171,034,255 \\ % 168,668,453
		% \hline
		% WMT newstest2014 & 3,003 & 72,435 \\
		% MSLT test & 3,543 & 31,338 \\
	\end{tabular}
	\caption{Statistics of French-English corpora.}
	\label{tab:data}
\end{table}

\paragraph{Data} We use the GYAFC corpus introduced by \citet{RaoT18} in all tasks. This corpus consists of informal sentences from Yahoo Answers paired with their formal rewrites by humans. The train split consists of 105K informal-formal sentence pairs whereas the dev/test sets consist of roughly 10K/5K pairs for both formality transfer directions, i.e., \texttt{I$\rightarrow$F} and \texttt{F$\rightarrow$I}.

We train MT systems on the concatenation of large diverse parallel corpora: (1) Europarl.v7 \citep{Koehn05}, which is extracted from the proceedings of the European Parliament, and tends to be more formal text; (2) News-Commentary.v14 \citep{BojarFFGHKM18}; (3) OpenSubtitles2016 \citep{LisonT16}, which consists of movie and television subtitles, covers a wider spectrum of styles, but overall tends to be informal since it primarily contains conversations. Following our previous work \citep{NiuRC18}, we use a bilingual semantic similarity detector to select 16M least divergent examples from $\sim$27.5M deduplicated sentence pairs in the original set \citep{VyasNC18}.

\paragraph{Preprocessing} We apply normalization, tokenization, true-casing, joint source-target BPE with 50,000 operations \citep{SennrichHB16a} and sentence-filtering (length 50 cutoff) to parallel training data. Table~\ref{tab:data} shows itemized translation data statistics after preprocessing.

\paragraph{Implementation Details} We build NMT models upon the attentional RNN encoder-decoder architecture \citep{BahdanauCB15} implemented in the Sockeye toolkit \citep{HieberDDVSCP17}. Our translation model uses a bi-directional encoder with a single LSTM layer of size 512, multilayer perceptron attention with a layer size of 512, and word representations of size 512. We apply layer normalization \citep{BaKH16}, add dropout to embeddings and RNNs \citep{GalG16} with probability 0.2, and tie the source and target embeddings as well as the output layer's weight matrix \citep{PressW17}. We train using the Adam optimizer \citep{KingmaB15} with a batch size of 64 sentences and we checkpoint the model every 1000 updates. The learning rate for baseline models is initialized to 0.001 and reduced by 30\% after 4 checkpoints without improvement of perplexity on the development set. Training stops after 10 checkpoints without improvement.

We build our FSMT models by fine-tuning the \baseMTL model with the dedicated synthetically supervised objectives described in Section \ref{sec:synthetic}, inheriting all settings except the learning rate which is re-initialized to 0.0001. The hyper-parameter $\alpha$ in Equation~\ref{eq:oti_obj} is set to 0.05.

\begin{table*}[t]
	\centering
	\begin{tabular}{l|rr|rr|rr|rr}
		Model & \texttt{I$\rightarrow$F} && \texttt{F$\rightarrow$I} && \texttt{I$\rightarrow$I} && \texttt{F$\rightarrow$F} & \\
		\hline
		None & 70.63 $\pm$ 0.23 && 37.00 $\pm$ 0.18 && 54.54 $\pm$ 0.44 && 58.98 $\pm$ 0.93 & \\
		\hline
		\textsc{Tag-Src} & 72.16 $\pm$ 0.34 & $\Delta$ & 37.67 $\pm$ 0.11 & $\Delta$ & 66.87 $\pm$ 0.58 & $\Delta$ & 78.78 $\pm$ 0.37 & $\Delta$ \\
		\textsc{Tag-Src-Block} & 72.00 $\pm$ 0.05 & -0.16 & 37.38 $\pm$ 0.12 & -0.29 & 65.46 $\pm$ 0.29 & \bftab -1.41 & 76.72 $\pm$ 0.39 & \bftab -2.06 \\
		\textsc{Tag-Src-Tgt} & 72.29 $\pm$ 0.23 & +0.13 & 37.62 $\pm$ 0.37 & -0.05 & 67.81 $\pm$ 0.41 & \bftab +0.94 & 79.34 $\pm$ 0.55 & \bftab +0.56 \\
	\end{tabular}
	\caption{BLEU scores for variants of side constraint in controlling style on all formality transfer and preservation directions. We report mean and standard deviation over five randomly seeded models. $\Delta$BLEU between each model and the widely used \textsc{Tag-Src} methods show that (1) blocking the visibility of source tags from the encoder (\textsc{Tag-Src-Block}) limits its formality control ability; (2) using style tags on both source and target sides (\textsc{Tag-Src-Tgt}) helps control formality better when considering the full range of formality change and formality preservation tasks.}
	\label{tab:tag}
\end{table*}

\section{Formality Control Evaluation}
\label{sec:tag}

Our goal is to determine a solid approach for formality control before adding synthetic supervision. For simplicity, we conduct this auxiliary evaluation of formality control on four sub-tasks that use monolingual style transfer data.

\paragraph{Tasks} Our task aims to test systems' ability to produce a formal or an informal paraphrase for a given English sentence of arbitrary style, as needed in FSMT. It is derived from formality transfer \citep{RaoT18}, where models transfer sentences from informal to formal (\texttt{I$\rightarrow$F}) or vice versa (\texttt{F$\rightarrow$I}). These sub-tasks only evaluate a model's ability in learning mappings between informal and formal languages. We additionally evaluate the ability of systems to preserve formality on informal to informal (\texttt{I$\rightarrow$I}) and formal to formal (\texttt{F$\rightarrow$F}) sub-tasks. GYAFC provides four reference target-style human rewrites for each source-style sentences in the test set. For formality preservation, the output is compared with the input sentence in the test set.

\paragraph{Models} All models are trained on bidirectional data, which is constructed by swapping the informal and formal sentences of the parallel GYAFC corpus and appending the swapped version to the original. The formality of each target sentence represents the desired input style.

We compare our approach, \textsc{Tag-Src-Tgt}, which attaches tags to both input and output sides, against two baselines. We first implement a baseline method which is trained only on the bidirectional data without showing the target formality (denoted as None). The second baseline is \textsc{Tag-Src}, the standard method that attaches tags to the source. In addition, we conduct an ablation study on the side constraint method using \textsc{Tag-Src-Block}, which attaches a tag to the source just like \textsc{Tag-Src} but blocks the visibility of the tag embeddings from the encoder and retains their connections to the decoder via the attention mechanism (Table~\ref{tab:tag}).

\paragraph{Results} Our approach, \textsc{Tag-Src-Tgt}, achieves the best performance overall, reaching the best BLEU scores for three of the four sub-tasks. Comparing with methods acknowledging the target formality (i.e., \textsc{Tag-Src*}), the None baseline gets slightly lower BLEU scores when it learns to flip the formality on \texttt{I$\rightarrow$F} and \texttt{F$\rightarrow$I} tasks.\footnote{As \citet{RaoT18} note, \texttt{F$\rightarrow$I} models yield lower BLEU than \texttt{I$\rightarrow$F} models because informal reference rewrites are highly divergent.} However, it performs much worse (10-20 BLEU points lower) on \texttt{I$\rightarrow$I} and \texttt{F$\rightarrow$F} tasks confirming that the None baseline is only able to flip formality and not to preserve it. The \textsc{Tag-Src} approach is able to preserve formality better than the None baseline, but not as well as \textsc{Tag-Src-Tgt}.

\textsc{Tag-Src-Block} lags behind \textsc{Tag-Src}, especially for formality preservation tasks (1-2 BLEU points lower). This discrepancy indicates that the attention mechanism only contributes a portion of the control ability. On the other hand, our proposed variant \textsc{Tag-Src-Tgt} performs better than \textsc{Tag-Src} on 3/4 tasks (i.e., \texttt{I$\rightarrow$F}, \texttt{I$\rightarrow$I}, and \texttt{F$\rightarrow$F}). 

Taken together, these observations show that the impact of tags is not limited to the attention model, and their embeddings influence the hidden representations of encoders and decoders positively. The auxiliary evaluation thus confirms that adding style tags to both source and target sequences is a good approach to model monolingual formality transfer, and therefore motivates using it in our FSMT models as well.

\section{Quantifying Differences Between Formal and Informal Outputs in FSMT}

Having established the effectiveness of our formality control mechanism, we now turn to the FSMT task and test whether synthetic supervision succeeds in introducing more differences between formal and informal outputs, regardless of translation quality. We will consider translation quality in the next section.

\paragraph{Tasks} We test FSMT approaches on two French-English translation test sets with diverse formality: WMT newstest2014\footnote{\url{http://www.statmt.org/wmt14/test-full.tgz}} and MSLT conversation test set\footnote{\url{https://www.microsoft.com/en-us/download/details.aspx?id=54689}}.
While each test set contains text of varying formality, the written language used in news stories is typically more formal than the spoken language used in conversations.

\paragraph{Baseline Models} We start with a standard \textbf{NMT} model which is trained with non-tagged French-English parallel data. This model achieves 28.63 BLEU on WMT and 47.83 BLEU on MSLT. We provide these BLEU scores for a sanity check on translation quality.\footnote{Detailed BLEU scores are available with released code.} FSMT models could receive up to two lower BLEU points on WMT and up to four lower BLEU points on MSLT. However, BLEU cannot be used to evaluate FSMT: given a single reference translation of unknown formality, BLEU penalizes both unwanted translation errors and correct formality rewrites. For example, given the reference ``we put together the new wardrobe", the good formal output ``we assembled the new wardrobe" and the incorrect output ``we accumulated the new wardrobe" would get the same BLEU score.

Next, we compare with \textbf{\baseMTL}, which performs zero-shot FSMT by training machine translation (MT) and formality transfer (FT) jointly. We also compare other two FSMT models introduced in our previous work \citep{NiuRC18} for completeness. (1) \textbf{\baseDS}. It performs data selection on MT training examples $(\vX,\vY)$ using CED in a standard way: it pre-trains language models for informal and formal English in the FT training data and calculates $\ced(\vY)=H_{informal}(\vY)-H_{formal}(\vY)$. We aim at using all parallel data, for fair comparison, we also conduct three-way tagging as introduced in Section~\ref{sec:synthetic}. An NMT model is then trained with the formality-tagged training pairs. (2) \textbf{\baseMTLDS}. It is the combination of \baseMTL and \baseDS and is trained on both tagged MT pairs and FT pairs. This method is similar to \methodEll in terms of tagging training examples using CED. However, \baseMTLDS uses standard offline language models while \methodEll can be interpreted as using source-conditioned online language models. %\footnote{We considered a pivoting approach (i.e., machine translation followed by formality transfer) in preliminary experiments, but it consistently underperforms multi-task baselines.}

\paragraph{Metrics} Since FSMT quality cannot be evaluated automatically, we devise an approach to quantify surface differences between formal and informal outputs of a given system to guide system development. We define the \textbf{Le}xical and \textbf{Po}sitional \textbf{D}ifferences (\textsc{LePoD}) score for this purpose, and will come back to FSMT evaluation using human judgments in the next section.

We first compute the pairwise Lexical Difference (\textsc{LeD}) based on the percentages of tokens that are not found in both outputs. Formally,
\begin{equation}\label{eq:led}
\textsc{LeD} = \frac{1}{2}\left(\frac{|S_1 \backslash S_2|}{|S_1|}+\frac{|S_2 \backslash S_1|}{|S_2|}\right),
\end{equation}
where $S_1$ and $S_2$ is a pair of sequences and $S_1 \backslash S_2$ indicates tokens appearing in $S_1$ but not in $S_2$.

\begin{figure}[t]
	\centering
	\includegraphics[width=\columnwidth]{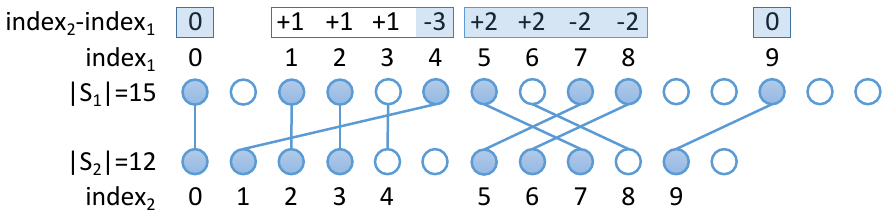}
	\caption{Comparing $S_1$ and $S_2$ with \textsc{LePoD}: hollow circles represent non-exact matched tokens, yielding a \textsc{LeD} score of $(\frac{7}{15}+\frac{4}{12})\times\frac{1}{2}=0.4$. Given the alignment illustrated above, the \textsc{PoD} score is $\frac{0+3+2+0}{10}=0.5$.}
	\label{fig:lepod}
\end{figure}

We then compute the pairwise Positional Difference (\textsc{PoD}). (1) We segment the sentence pairs into the longest sequence of phrasal units that are consistent with the word alignments. Word alignments are obtained using the latest METEOR software \citep{DenkowskiL14}, which supports stem, synonym and paraphrase matches in addition to exact matches. (2) We compute the maximum distortion within each segment. To do these, we first re-index $N$ aligned words and calculate distortions as the position differences (i.e., $\text{index}_2\,\text{-}\,\text{index}_1$ in Figure \ref{fig:lepod}). Then, we keep a running total of the distortion array $(d_1, d_2, \dots)$, and do segmentation $p=(d_i,\dots,d_j)\in P$ whenever the accumulation is zero (i.e., $\sum p=0$). Now we can define
\begin{equation}\label{eq:pod}
\textsc{PoD} = \frac{1}{N}\sum_{p\in P}\max(\abs(p)).
\end{equation}
In extreme cases, when the first word in $S_1$ is reordered to the last position in $S_2$, \textsc{PoD} score approaches 1. When words are aligned without any reordering, each alignment constitutes a segment and \textsc{PoD} equals 0.

\begin{table}[t] 
	\centering
	\tabcolsep=0.1cm
	\begin{tabular}{lrrlrr}
		\toprule
		& \multicolumn{2}{c}{WMT} & \phantom{a} &\multicolumn{2}{c}{MSLT} \\
		\cmidrule{2-3} \cmidrule{5-6}
		& \textsc{LeD} & \textsc{PoD} & & \textsc{LeD} & \textsc{PoD} \\
		\midrule
		NMT & 0 & 0 & & 0 & 0 \\
		\addlinespace[0.25em]
		\multicolumn{3}{l}{\textit{FSMT Baselines}}\\
		\baseDS & 9.27 & 6.44 & & 8.18 & 1.10 \\
		\baseMTL & 10.89 & 7.76 & & 11.97 & 1.41 \\
		\baseMTLDS & 11.51 & 8.35 & & 10.29 & 1.54 \\
		\addlinespace[0.25em]
		\multicolumn{4}{l}{\textit{Multi-Task w/ Synthetic Supervision}}\\
		Synth. Target & 10.97 & 7.25 & & 12.40 & 1.63 \\
		Synth. Style & \bftab 14.53 & \bftab 12.58 & & \bftab 14.52 & \bftab 2.19 \\
		\bottomrule
	\end{tabular}
	\caption{\textsc{LePoD} scores (percentages) show that synthetic supervision introduces more changes between formal and informal outputs than baselines. \methodEll (OSI) produces the most diverse informal/formal translations.}
	\label{tab:fsmt-auto}
\end{table}

\paragraph{Findings} Multi-task methods introduce more differences between formal and informal translations than NMT baselines, and synthetic supervision with \methodY obtains the best lexical and positional difference scores overall (Table~\ref{tab:fsmt-auto}). Specifically, \baseMTL and \baseMTLDS get similar lexical and positional variability, and both surpass \baseDS. \methodY has much larger positional discrepancy scores than all other methods, which indicates that it produces more structural diverse sentences. However, larger surface changes are more likely to alter meaning, and the changes are not guaranteed to be formality-oriented. We therefore turn to human judgments to assess whether meaning is preserved, and whether surface differences are indeed formality related.

\section{Human Evaluation of FSMT}
\label{sec:human}

\begin{figure*}[t]
	\centering
	\subfloat[Formality of informal translations]{
		\label{fig:human-formality-2I}
		\includegraphics[width=0.32\textwidth]{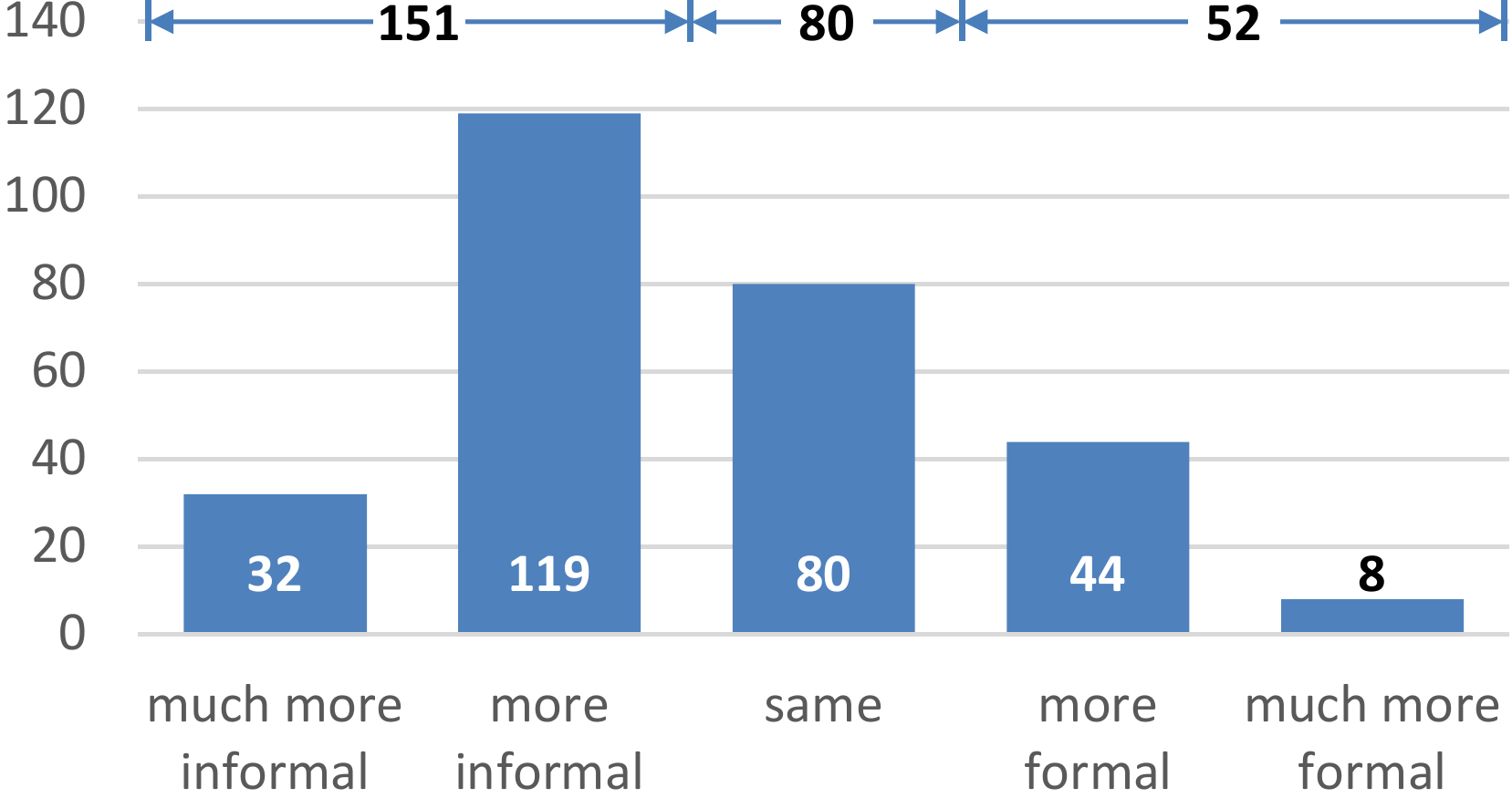}
	}
	\subfloat[Formality of formal translations]{
		\label{fig:human-formality-2F}
		\includegraphics[width=0.32\textwidth]{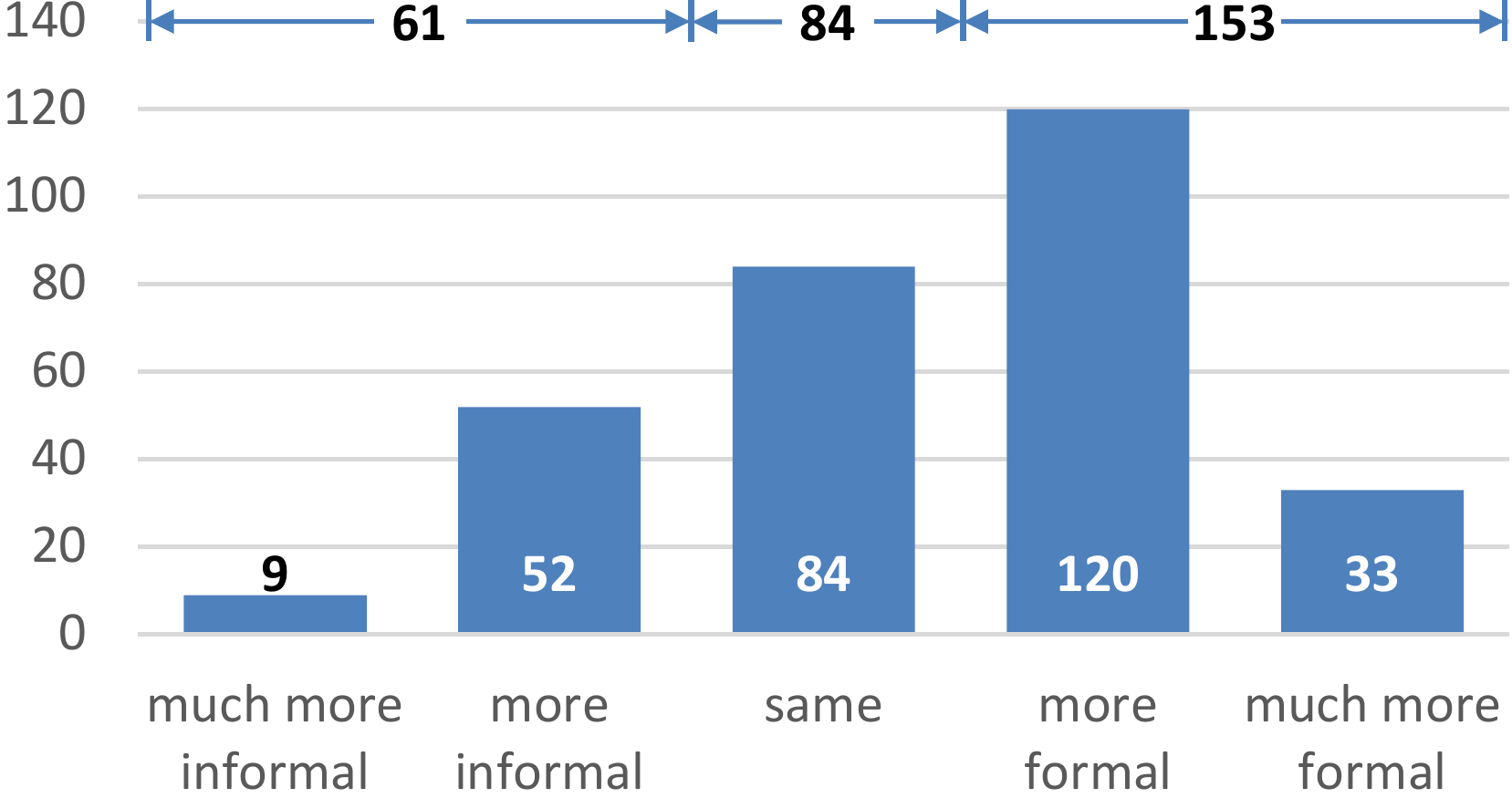}
	}
	\subfloat[Meaning Preservation]{
		\label{fig:human-meaning}
		\includegraphics[width=0.32\textwidth]{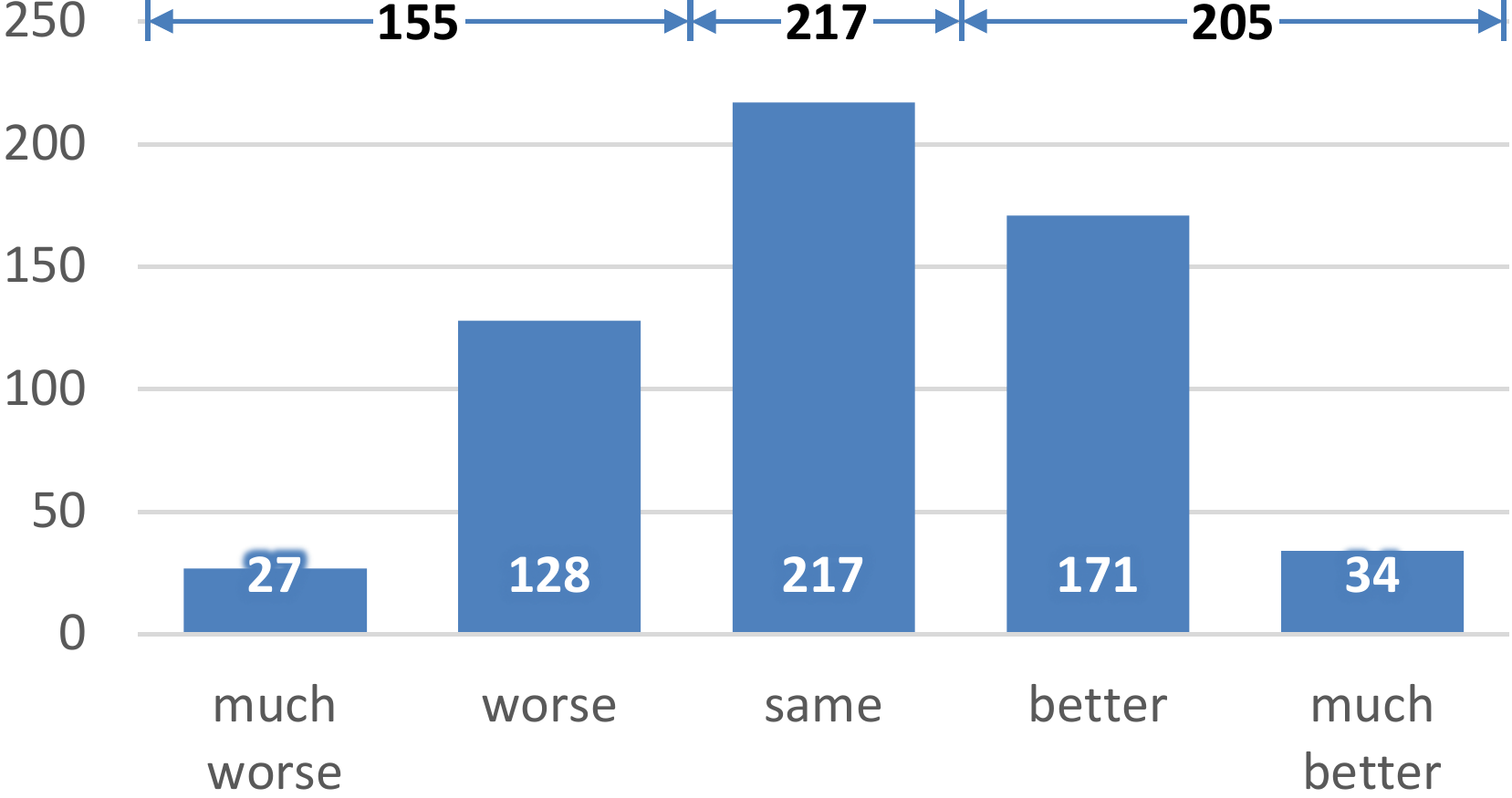}
	}
	\caption{Win/Tie/Loss counts when comparing \methodEll to \baseMTL. Informal translations generated by OSI are annotated as more informal than \baseMTL, while formal translations are annotated as more formal. The OSI model also gets more instances that better preserve the meaning.}
	\label{fig:fsmt-human}
\end{figure*}

\begin{table*}[t] 
	\centering
	\begin{tabular}{l|c|cc|ccc|c}
		Model & identical & contr. & filler & quot. & poss. & y/n & $\Delta$length \\
		\hline
		\baseMTL & 2,140 (33\%) & \hspace{9pt}915 & 530 & 146 & 46 & 13 & 1.30 \\
		\methodY & 1,868 (29\%) & \bftab 1,370 & \bftab 635 & 145 & 41 & 21 & 1.58 \\
		\methodEll & \bftab 1,385 (21\%) & 1,347 & 530 & \bftab 252 & \bftab 86 & \bftab 33 & \bftab 4.57 \\
	\end{tabular}
	\caption{Heuristic analysis of the differences between informal and formal translations. Synthetic supervision introduce more changes. \methodY usually performs simple substitutions while \methodEll performs more less-deterministic changes. \methodEll generates more complete and longer formal translations.}
	\label{tab:breakdown}
\end{table*}

Evaluating FSMT systems requires evaluating whether their outputs correctly convey the meaning of the source, and whether the differences between their formal and informal outputs are indicative of formality. Neither LePoD nor BLEU can assess these criteria automatically. We therefore conduct human evaluation to investigate whether synthetic supervision improves over our reimplementation of the state-of-the-art approach (\baseMTL).

\paragraph{Methodology} Following \citet{RaoT18} and our previous work \citep{NiuRC18}, we adopt the following evaluation criteria: \textit{meaning preservation} and \textit{formality difference}.\footnote{We do not evaluate fluency because both \citet{RaoT18} and \citet{NiuRC18} show various automatic systems achieve an almost identical fluency level. Annotators also have systematically biased feeling in fluency when comparing formal and informal sentences \citep{NiuMC17,RaoT18}.} Our evaluation scheme asks annotators to directly compare sentence pairs on these two criteria and obtains win:tie:loss ratios.
\begin{description}
	\item[Meaning Preservation] We ask annotators to compare outputs of two systems against the reference translation, and decide which one better preserves the reference meaning.
	\item[Formality Difference] We ask annotators to compare outputs of two systems and decide which is more formal.
\end{description}

We randomly sample $\sim$150 examples from WMT and MSLT respectively, and obtain judgments for informal and formal translations of each example. We collect these judgments from 30 volunteers who are native or near-native English speakers. Annotators only compare translations of the same (intended) formality generated by different systems. Identical translation pairs are excluded. Each comparison receives five independent judgments, unless the first three judgments are identical.

The inter-rater agreement using Krippendorff's alpha is $\sim$0.5.\footnote{In a sentential formality scoring task, \citet{PavlickT16} also report relatively low inter-annotator agreement with other measurements.} It indicates that there is some variation in annotators' assessment of language formality. We therefore aggregate independent judgments using MACE \citep{HovyBVH13}, which estimates the competence of annotators.

\paragraph{\bf Findings} Overall, the human evaluation shows that synthetic supervision successfully improves desired formality of the output while preserving translation quality, compared to the multi-task baseline, which represents prior state-of-the-art \citep{NiuRC18}. Figure~\ref{fig:human-formality-2I} and \ref{fig:human-formality-2F} show that \methodEll generates informal translations that are annotated as more informal  (\textbf{win:tie:loss=151:80:52}), while formal translations are annotated as more formal (\textbf{win:tie:loss=153:84:61}). For both cases, the win-loss differences are significant with $p<0.001$ using the sign test, where ties are evenly distributed to wins and losses as suggested by \citet{Demsar06}. The results confirm that synthetic supervision lets the model better tailor its outputs to the desired formality, and suggest that the differences between formal and informal outputs detected by the \textsc{LePoD} scores are indeed representative of formality changes. Figure~\ref{fig:human-meaning} shows that \methodEll preserves the meaning of the source better than \baseMTL (\textbf{win:tie:loss=205:217:155}). The win-loss difference for meaning preservation is still significant with $p<0.02$, but is less strong than formality difference.

\section{Analysis}

\begin{table*}[t]
	\centering
	\resizebox{\textwidth}{!}{
	\begin{tabular}{lll}
		\bf Type & \bf Informal translation & \bf Formal translation \\
		\hline
		Filler & \textbf{And} I think his wife has family there. & I think his wife has family there. \\
		\hline
		Completeness $\blacktriangledown$ & & \\
		Quotation & The gas tax is simply not sustainable, said Lee. & \textbf{``}The gas tax is simply not sustainable,\textbf{"} said Lee. \\
		Yes-No & You like shopping? & \textbf{Do} you like shopping? \\
		Subject & Sorry it's my fault. & \textbf{I'm} sorry it's my fault. \\
		Article & Cookies where I work. & \textbf{The} cookies where I work. \\
		Relativizer & Other stores you can't buy. & The other stores \textbf{where} you can't buy. \\
		% Politeness & Can you come help me? & Can you come and help me please? \\
		\hline
		Paraphrasing $\blacktriangledown$ & & \\
		Contraction & I think \textbf{he'd} like that, but \textbf{we'll} see. & I think \textbf{he would} like that, but \textbf{we will} see. \\
		Possessive & \textbf{Fay's innovation} perpetuated over the years. & \textbf{The innovation of Fay} has perpetuated over the years. \\
		Adverb & I \textbf{told you already}. & I \textbf{already told you}. \\
		Idiom & Hi, \textbf{how's it going}? & Hi, \textbf{how are you}? \\
		Slang & You \textbf{gotta} let him digest. & You \textbf{have to} let him digest. \\
		Word-1 & Actually my \textbf{dad}'s some kind of technician & In fact, my \textbf{father} is some kind of technician \\
		& so he understands, but my \textbf{mom}'s very old. & so he understands, but my \textbf{mother} is very old. \\
		Word-2 & \textbf{Maybe} a little more in \textbf{some} areas. & \textbf{Perhaps} a little more in \textbf{certain} areas. \\
		Word-3 & It's \textbf{really necessary} for our nation. & This is \textbf{essential} for our nation. \\
		Phrase-1 & Yeah, \textbf{me neither}. & Yeah, \textbf{neither do I}. \\
		Phrase-2 & I think he's moving to California \textbf{now}. & I think he is moving to California \textbf{at the moment}. \\
		Phrase-3 & It could be \textbf{a Midwest thing}. & This could be \textbf{one thing from the Midwest}. \\
		\hline
	\end{tabular}}
	\caption{Range of differences between informal and formal translations from the \methodEll model output.}
	\label{tab:breakdown-examples}
\end{table*}

How do informal and formal translations differ from each other? Manual inspection reveals that most types of changes made by human rewriters \citep{PavlickT16,RaoT18}, including use of filler words, completeness of output and various types of paraphrasing, are observed in our system outputs (see examples in Table~\ref{tab:breakdown-examples}). We quantify such changes further semi-automatically.

We first check how often formal and informal translations are identical. This happens less frequently with synthetic supervision (Table~\ref{tab:breakdown}) than with the baseline multi-task system: \methodEll system introduces changes between formal and informal translations 12\% more often in 6,546 test examples compared to the baseline.

Then, we use rules to check how often simple formality change patterns are found in FSMT outputs (Table~\ref{tab:breakdown}). A sentence can be made more formal by expanding contractions (contr.) and removing unnecessary fillers such as conjunctions (\textit{so/and/but}) and interjections (\textit{well}) at the beginning of a sentence (filler). \methodY performs these changes more frequently. We also examine the introduction of quotation marks in formal translations (quot.); using possessive \textit{of} instead of possessive \textit{'s} (poss.); and rewrites of informal use of declarative form for yes-no questions (y/n). \methodEll output matches these patterns better than other systems.

Next, we conduct a manual analysis to understand the nature of remaining differences between formal and informal translations of \methodEll. We observe that ellipsis is frequent in informal outputs, while formal sentences are more complete, using complement subjects, proper articles, conjunctions, relative pronouns, etc. This is reflected in their longer length ($\Delta$length in Table~\ref{tab:breakdown} is the average length difference in characters). Lexical or phrasal paraphrases are frequently used to convey formality, substituting familiar terms with more formal variants (e.g., ``grandma'' vs. ``grandmother''). Examining translations with large \textsc{PoD} scores shows that \methodEll is more likely to reorder adverbs based on formality: e.g., ``I told you already"~(\texttt{I}) vs. ``I already told you"~(\texttt{F}).

A few types of human rewrites categorized by \citet{PavlickT16} and \citet{RaoT18} are not observed here. For example, our models almost always produce words with correct casing and standard spelling for both informal and formal languages. This matches the characteristics of the translation data we used for training.

Finally, we manually inspect system outputs that fail to preserve the source meaning and reveal some limitations of using synthetic supervision. (1) Inaccurate synthetic labels introduce noise. \methodY sometimes generates ``I am not sure" as the formal translation, regardless of the source. We hypothesize that this is due to the imperfect synthetic translations generated by the formality transfer sub-model reinforce this error pattern. (2) Synthetic data may not reflect the true distribution. Occasionally, \methodEll drops the first word in a formal sentence even if it is not a filler, e.g. ``\st{On} Thursday, ..."  We hypothesize that labeling too many formal/informal examples of similar patterns could lead to ignoring context. While \methodEll improves meaning preservation comparatively, it still bears the challenge of altering meaning when fitting to a certain formality, such as generating ``there will be no longer than the hill of Runyonyi" when the reference is ``then only Rumyoni hill will be left".

\section{Related Work}

Controlling the output style in MT has received sparse attention. The pioneering work by \citet{MimaFI97} improves rule-based MT using extra-linguistic information such as speaker's role and gender. With the success of statistical MT models, people usually define styles by selecting representative data. After pre-selecting relevant data offline, \citet{LewisFX15} and \citet{WeesBM16} build conversational MT systems, \citet{RabinovichMPSW17} and \citet{MichelN18} build personalized (gender-specific) MT systems, \citet{SennrichHB16} control the output preference of T-V pronouns, while \citet{YamagishiKSK16} control the active/passive voice of the translation. In contrast, we dynamically generate synthetic supervision and our methods outperform offline data selection.

Multi-task FSMT is closely related to zero-shot multilingual NMT. \citet{JohnsonSLKWCTVW17} first built a multilingual NMT system using shared NMT encoder-decoders for all languages with target language specifiers. The resulting system can translate between language pairs that are never trained on, but performs worse than supervised models and even the simple pivoting approach for those language pairs. Strategies to mitigate this problem include target word filtering \citep{HaNW17}, dedicated attention modules \citep{BlackwoodBW18}, generating dedicated encoder-decoder parameters \citep{PlataniosSNM18} and encouraging the model to use source-language invariant representations \citep{ArivazhaganBFAJM19}. We address this problem from a different perspective for the FSMT task by automatically inferring style labels. Our \methodY approach is similar in spirit to a contemporaneous method that encourages the model to produce equivalent translations of parallel sentences into an auxiliary language \citep{Al-ShedivatP19}.

\section{Conclusion}

This paper showed that synthetic supervision improves multi-task models for formality-sensitive machine translation. We introduced a novel training scheme for multi-task models that, given bilingual parallel examples and monolingual formality transfer examples, automatically generate synthetic training triples by inferring the target formality from a given translation pair. Human evaluation shows that this approach outperforms a strong multi-task baseline by producing translations that better match desired formality levels while preserving the source meaning. Additional automatic evaluation shows that (1) attaching style tags to both input and output sequences improves the ability of a single model to control formality, by not only transferring but also preserving formality when required; and (2) synthetic supervision via \methodY introduces more changes between formal and informal translations of the same input. Analysis shows that these changes span almost all types of changes made by human rewriters.

Taken together, these results show the promise of synthetic supervision for style-controlled language generation applications. In future work, we will investigate scenarios where style transfer examples are not readily available, including for languages other than English, and for style distinctions that are more implicit and not limited to binary formal-informal distinctions.

\small
%\fontsize{9.0pt}{10.0pt} \selectfont
\bibliography{fsmt}
\bibliographystyle{aaai}

\appendix

\section{Details of the Human Evaluation}

As described in the main paper, we assess model outputs on two criteria: \textit{meaning preservation} and \textit{formality difference}.
\begin{description}
	\item[Meaning Preservation] The following instruction is provided to annotators.
\end{description}

\begin{quotation}\sffamily
	For each task, you will be presented with an English sentence and two rewrites of that sentence. Your task is to judge which rewrite better preserves the meaning of the original and choose from:
	\begin{itemize}
		\item Rewrite 1 is much better
		\item Rewrite 1 is better
		\item No preference between Rewrite 1 and Rewrite 2 (no difference in meaning or hard to say)
		\item Rewrite 2 is better
		\item Rewrite 2 is much better
	\end{itemize}
	
	Note that this task focuses on differences in content, so differences in style (such as formality) between the original and rewrites are considered okay. [Some examples with explanations are provided.]
\end{quotation}

\begin{description}
	\item[Formality Difference] The following instruction is provided to annotators.
\end{description}

\begin{quotation}\sffamily
	People use different varieties of language depending on the situation: formal language is required in news articles, official speeches or academic assignments, while informal language is more appropriate in instant messages or spoken conversations between friends.
	
	You will be presented with two English sentences, and your task is to decide which one is more formal and choose from:
	\begin{itemize}
		\item Sentence 1 is much more formal
		\item Sentence 1 is more formal
		\item No preference between Sentence 1 and Sentence 2 (no difference in formality or hard to say)
		\item Sentence 2 is more formal
		\item Sentence 2 is much more formal
	\end{itemize}
	
	Keep in mind:
	\begin{itemize}
		\item Language formality can be affected by many factors, such as the choices of grammar, vocabulary, and punctuation.
		\item The sentences in the pair could have different meanings. Please rate the formality of the sentences independent of their meaning.
		\item The sentences in the pair could be nonsensical. Please rate the formality of the sentences independent of their quality.
	\end{itemize}
	
	Generally, a sentence with small formality changes such as fewer contractions, proper punctuation or some formal terms is considered ``more formal". A sentence is considered ``much more formal" if it contains multiple indicators of formality, or if the sentence construction itself reflects a more formal style. That said, feel free to use your own judgment for doing the task if what you see is not covered by these examples. [Some examples with explanations are provided.]
\end{quotation}

\section{Extended Formality Control Evaluation}

\begin{table*}[t]
	\begin{center}
		\begin{tabular}{l|rr|rr|rr|rr}
			Model & \texttt{I$\rightarrow$F} && \texttt{F$\rightarrow$I} && \texttt{I$\rightarrow$I} && \texttt{F$\rightarrow$F} & \\
			\hline
			\textsc{Tag-Src-Tgt} & 72.29 $\pm$ 0.23 & $\Delta$ & 37.62 $\pm$ 0.37 & $\Delta$ & 67.81 $\pm$ 0.41 & $\Delta$ & 79.34 $\pm$ 0.55 & $\Delta$ \\
			\textsc{Factor-Concat} & 72.47 $\pm$ 0.11 & +0.18 & 37.62 $\pm$ 0.26 & 0.00 & 67.03 $\pm$ 0.36 & -0.78 & 79.80 $\pm$ 0.38 & +0.46 \\
			\textsc{Factor-Sum} & 72.43 $\pm$ 0.29 & +0.14 & 37.78 $\pm$ 0.26 & +0.16 & 67.24 $\pm$ 0.56 & -0.57 & 80.34 $\pm$ 0.46 & +1.00 \\
			\textsc{Pred-Concat} & 72.35 $\pm$ 0.16 & +0.06 &  37.62 $\pm$ 0.13 & 0.00 & 66.69 $\pm$ 0.21 & -1.12 & 77.85 $\pm$ 0.31 & -1.49 \\
			\textsc{Pred-Sum} & 72.02 $\pm$ 0.30 & -0.27 & 37.41 $\pm$ 0.17 & -0.21 & 66.15 $\pm$ 0.41 & -1.66 & 77.62 $\pm$ 0.28 & -1.72 \\
			\textsc{BOS} & 72.08 $\pm$ 0.22 & -0.21 & 37.56 $\pm$ 0.13 & -0.06 & 66.40 $\pm$ 0.23 & -1.41 & 77.43 $\pm$ 0.34 & -1.91 \\
			\textsc{Bias} & 71.58 $\pm$ 0.31 & -0.71 & 37.52 $\pm$ 0.15 & -0.10 & 63.66 $\pm$ 0.51 & -4.15 & 73.24 $\pm$ 0.55 & -6.10 \\
		\end{tabular}
	\end{center}
	\caption{BLEU scores of various methods for controlling the style on four formality transfer (preservation) directions. The numbers before and after `$\pm$' are the mean and standard deviation over five randomly seeded models. Methods are compared with \textsc{Tag-Src-Tgt} and the $\Delta$BLEU scores are listed.}
	\label{tab:transfer}
\end{table*}

While the implementations of neural language generation converge to an encoder-decoder framework, the design choice of controlling the style is full of variety. The style information could be injected into different parts of the heterogeneous neural network and all roads lead to Rome. However, those implementations have never been compared with a controlled experiment and analyzed contrastively. We therefore conduct a benchmark test on a four-way formality rewriting task introduced in the main paper.

In order to focus on the designing of style-sensitive neural models, we compare methods performing formality rewriting with a single encoder and a single decoder, in contrast with a dedicated model or decoder for each transfer direction \cite{RaoT18,FuTPZY18}. The attention mechanism is the \textit{de facto} standard for language generation \cite{BahdanauCB15,LuongPM15,XuBKCCSZB15}, we therefore compare methods being compatible with the attention mechanism, in contract with methods that compress the content into one single vector \cite{MuellerGJ17,HuYLSX17,ShenLBJ17,FuTPZY18}.

We compare the following methods, with a focus on their complexity and effectiveness.
\begin{description}
	\item[\textsc{Tag-Src-Tgt}] This is the method used in the main paper. It attaches style tags to both source and target sequences. Each additional tag occupies one embedding vector, which has a size of $O(E_w)$, where $E_w$ is the word embedding size.
	\item[\textsc{Factor}] The style information can be incorporated as source word factors, which is implemented as style factor embeddings concatenated to the word embeddings \citep{SennrichH16}, i.e., $\tilde{\vX}_i=[\vX_i;\vX^\texttt{style}_i]$. \citet{KorotkovaDF18} adopt this design choice for multiple-style transfer due to its flexibility. Summing the factor and word embeddings of the same size is another combination strategy and we name it \textsc{Factor-Sum}, which uses $O(E_w)$ space, as opposite to \textsc{Factor-Concat}, which uses $O(E_s)$ space. $E_s$ is the style embedding size and usually much smaller than $E_w$.\footnote{We use $E_s=5$ in our experiments.}
	\item[\textsc{Pred}] Alternatively, we can inject the style information later on to the decoder by concatenating style embeddings to predicted target word embeddings, i.e., $\tilde{\vY}_t=[\vY_t;\vY^\texttt{style}_t]$. \citet{FiclerG17} use this method in the style-conditioned language generation. Summing can also be used here, and we name these two variants \textsc{Pred-Concat} and \textsc{Pred-Sum}. The space complexities are $O(E_s)$ and $O(E_w)$.
	
	Note that for both \textsc{Factor} and \textsc{Pred}, the computational complexity also increase proportionally with the sequence length since the style embeddings are combined with word embeddings for each time step.
	\item[\textsc{BOS}] Analogical to \textsc{Tag}, the target style embeddings can be dynamically attached to the target sequence as a begin-of-sequence symbol (\texttt{<BOS>}). This approach has been successfully applied to multiple-attribute text rewriting \citep{LampleSSDRB19}. Each stylistic \texttt{<BOS>} embedding occupies $O(E_w)$ space.
	\item[\textsc{Bias}] The bias parameter influences the model's lexical choice in the output layer (i.e. $\softmax(\vW\vh_t+\vb)$), so we can assign a dedicated bias for each style. \citet{MichelN18} use this technique in personalized NMT. Each dedicated bias has a size of $O(V)$, where $V$ is the vocabulary size.
\end{description}

We compare all methods to \textsc{Tag-Src-Tgt}, which is introduced in the main paper. Experimental settings and implementation details are identical to the intrinsic evaluation in the main paper and we report scores in Table~\ref{tab:transfer}.

The other method family incorporating the style information as early as at the encoding stage, \textsc{Factor}, waxes and wanes, and performs similar to \textsc{Tag-Src-Tgt}. Remaining methods that incorporate the style information only to the decoder, on the other hand, get lower BLEU scores across the board.

\end{document}